# Are You an Introvert or Extrovert? Accurate Classification With Only Ten Predictors

Chaehan So
Information & Interaction Design
Humanities, Arts & Social Sciences Division, Yonsei University
Seoul, South Korea
Email: cso@yonsei.ac.kr

*Abstract*— This paper investigates how accurately the prediction of being an introvert vs. extrovert can be made with less than ten predictors. The study is based on a previous data collection of 7161 respondents of a survey on 91 personality and 3 demographic items.

The results show that it is possible to effectively reduce the size of this measurement instrument from 94 to 10 features with a performance loss of only 1%, achieving an accuracy of 73.81% on unseen data. Class imbalance correction methods like SMOTE or ADASYN showed considerable improvement on the validation set but only minor performance improvement on the testing set.

*Keyword*s—personality traits, extraversion, introversion, class imbalance, variable importance.

## I. INTRODUCTION

In psychology, the differences between people's personality is often described on the trait level [1]. An abundance of research has provided evidence for such inter-individual differences that are stable over time. One prominent example is the so-called big five model by McCrae and John [2] that postulates the existence of five independent personality traits: conscientiousness, openness, extraversion, agreeableness and neuroticism. The big five model has been confirmed by numerous validation studies, e.g. conscientiousness ($\beta = .37$) and extraversion ($\beta = .23$) predict subjective well-being [3], and extraversion is negatively correlated with self-consciousness ($r = -.58$, $p < .05$) [4]. Many of these studies show the elevated relevance of extraversion as a predictor of many psychological phenomena.

Extraversion is often described on a bipolar continuum where low scores on extraversion are categorized as introversion [5]. Psychometrics, the science of measuring psychological dimensions, defines that personality traits can be measured as latent variables which can be captured by scales with sufficiently high internal consistency [6].

The relevance of identifying extraversion correctly becomes apparent by two complementary findings: On the one side, extraversion originates from a dopaminergic circuit involving the ventral striatum [7]. This fact may be responsible for the positive relationship between extraversion and subjective well-being consistently found in happiness research. On the other side, less extraverted people can induce positive affect by acting extraverted, i.e. by exhibiting extraverted behavior [8]. This may represent an effective measure for introverts who suffer from the negative aspects of introverted behavior like loneliness that leads to more negative affect [9].

## II. METHOD

### A. Multidimensional Introversion-Extroversion Scales Dataset

The open data movement has long entered the psychological field and led to various efforts to make interesting psychological datasets publicly available. One such offering is the *Open-Source Psychometric Project* [10] that collects data on a large scale on various personality assessment measurement scales. Among them, the *Multidimensional Introversion-Extraversion Scales* (MIES) dataset was released on 19 Aug 2019 and contains 7188 questionnaire responses.

The MIES dataset consists of the target variable *IE*, i.e. the self-identification as introvert or extrovert, and 95 features, thereof 91 items measuring the degree of extraversion on a 5-point Likert scale (from 1 = *Disagree* to 5 = *Agree*), 3 demographic items (country, gender, English native speaker), 1 item indicating the date of survey submittal. The latter item was removed as it didn't represent any predictive value for the introvert/extrovert self-identification, leading to 94 features.

Participants were 58.1% female and 37.8% male, to a high proportion (68.9%) English native speaker, and spread over 121 countries, mainly *US* (47.6 %), *Great Britain* (7.16 %), *Canada* (6.26 %), *Australia* (4.50 %), *Germany* (2.70 %), India (2.07 %), Philippines (1.76 %), Brazil(1.27 %), Poland (1.27 %), Romania (1.08 %), Malaysia (1.08 %), Netherlands (1.03 %), New Zealand (0.96 %), France (0.95 %) and Sweden (0.91 %).

### B. Technological Infrastructure

The data analysis was performed on a virtual machine with Rstudio Server running on 96 CPU cores (Intel Skylake), and 250GB RAM in the Google Cloud Compute Engine.

## C. Data Preprocessing

The dataset was analyzed for missing values. 27 observations did not contain any value in the target variable ("IE" = introvert-extrovert self-identification), and thus were removed from the dataset. This removal led to a final sample size of 7161. In addition to the personality items, two technical items denoted the item position in the survey and the time elapsed for this item in milliseconds. These technical items were removed as they didn't contain any predictive information.

TABLE 1: BENCHMARKING ACCURACY RESULTS – BASELINE FEATURE SET

| Rank | Variable Importance | Item | Item Wording |
|---|---|---|---|
| 1 | 100 | Q83A | I keep in the background. |
| 2 | 92.412 | Q91A | I talk to a lot of different people at parties. |
| 3 | 54.422 | Q82A | I don't talk a lot. |
| 4 | 28.231 | Q80A | I love large parties. |
| 5 | 25.257 | Q90A | I start conversations. |
| 6 | 16.802 | Q81A | I am quiet around strangers. |
| 7 | 14.500 | Q10A | I prefer to socialize 1 on 1, than with a group. |
| 8 | 14.039 | Q84A | I don't like to draw attention to myself. |
| 9 | 13.611 | Q14A | I want a huge social circle. |
| 10 | 12.672 | Q7A | I spend hours alone with my hobbies. |

*Note: random forests with 10-fold cross-validation in 2 repetitions*

## D. Feature Selection

The full dataset (n = 7161) containing 94 features was estimated for the classification of the target variable by the random forests algorithm [11] using 10-fold cross-validation in 2 repetitions. From this estimation, the variable importance for all features were derived. Table 1 shows the relative influence on the prediction normalized by 100 for the best predictor (*Q83A: "I keep in the background."*).

The accuracy on the hold-out validation set was 74.49% and served as baseline for the subsequent benchmarking on the reduced dataset.

## E. Counterbalancing Class Imbalance

The target variable contained three classes that were distributed unevenly. The majority class (introverted) represented 61.5% of the target variable. Two methods were applied to counterbalance the encountered class imbalance.

One of the most commonly applied methods is the synthetic minority oversampling technique (SMOTE) [12]. This method synthesizes new class instances of the minority class by an interpolation with k-nearest neighbors. The newly generated minority class samples serve to reduce the bias during training. Another class imbalance correcting method is adaptive synthesis sampling (ADASYN) [13]. This method was conceived to correct the minority class by oversampling but also adaptively give higher weights for the misclassified samples.

## III. RESULTS

### A. Benchmarking on Reduced Dataset

The dataset was substantially reduced from 94 features to the top 10 predictors of the preceding variable importance analysis. This benchmarking was run again with 10-fold cross-validation, but with 10 instead of 2 repetitions. The dataset was split with a 85:15 split ratio amounting to 6087 samples in the training set and 1074 samples in the testing set.

To achieve the same data distribution, the data split was performed with stratification, maintaining the same relative frequencies for each class level. Table 2 displays the five machine learning algorithms which were selected for the benchmarking.

TABLE 2: BENCHMARKING ALGORITHMS

| Nr | Reference | Algorithm Name | R Package |
|---|---|---|---|
| 1 | [14] | k-nearest neighbors | knn |
| 2 | [15] | Single-layer neural network | nnet |
| 3 | [16] | Gradient boosting machines | gbm |
| 4 | [11] | Random forests | randomForest |
| 5 | [17] | Support vector machines | kernlab |

Table 3 and Figure 1 display the results of the first benchmarking run, sorted by testing set performance. These results are interesting in the following aspects:

- The best algorithms achieve accuracy values on the cross-validation hold-out set (between 73.7 - 73.8%) that reach approximately 99.0% of the baseline performance (74.49%).

- This performance can be seen as a relatively good result considering it is distinctly higher than the *no information rate* (61.51%).

- The accuracy between cross-validation and testing set does not differ dramatically. This result is expected due to data stratification and because cross-validation uses a hold-out set to estimate accuracy that should resemble the testing set if no tuning was performed.

- The algorithms do not differ distinctly in testing set performance, yet random forests perform weakly.

TABLE 3: BENCHMARKING RESULTS – ORIGINAL DATASET

| algorithm | Cross-valid Acc – mean | Cross-valid Acc – sd | Cross-valid Kappa – mean | Cross-valid Kappa – sd | Testingset Accuracy |
|---|---|---|---|---|---|
| gbm | 73.83% | 1.45% | 48.80% | 2.94% | 73.81% |
| nnet | 73.71% | 1.44% | 48.77% | 2.91% | 73.63% |
| knn | 71.69% | 1.62% | 44.54% | 3.14% | 72.04% |
| rf | 73.27% | 1.56% | 47.38% | 3.05% | 71.95% |
| svmRadial | 73.62% | 1.56% | 47.07% | 3.31% | 71.95% |

*Note: Accuracy on 10-fold cross-validation in 10 repetitions, train/test split 85:15*

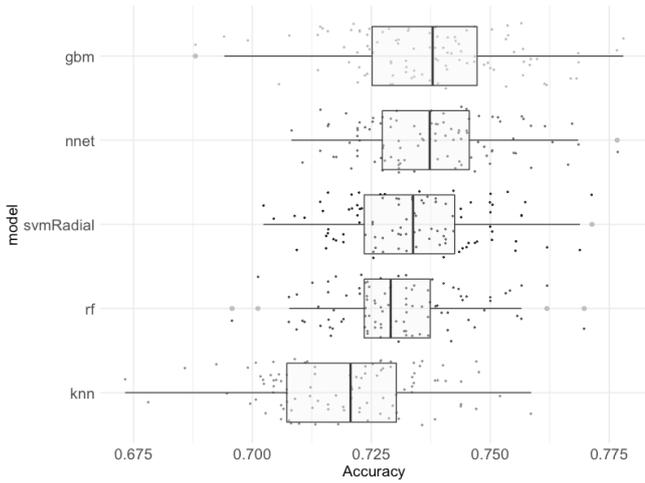

Figure 1: Benchmarking Results – Holdout Set – Original Dataset

### B. Benchmarking on SMOTE Dataset

Table 4 displays the results of the second benchmarking run, performed on the dataset corrected by SMOTE.

TABLE 4: BENCHMARKING RESULTS – SMOTE DATASET

| algorithm | Cross-valid Acc – mean | Cross-valid Acc – sd | Cross-valid Kappa – mean | Cross-valid Kappa – sd | Testingset Accuracy |
|---|---|---|---|---|---|
| gbm | 84.23% | 1.00% | 73.60% | 1.64% | 72.79% |
| rf | 84.34% | 0.81% | 73.52% | 1.37% | 72.79% |
| svmRadial | 79.76% | 0.78% | 64.53% | 1.37% | 69.80% |
| nnet | 78.83% | 0.92% | 63.02% | 1.61% | 68.97% |
| knn | 80.43% | 0.88% | 65.86% | 1.56% | 66.08% |

*Note: Accuracy on 10-fold cross-validation in 10 repetitions, train/test split 85:15*

The SMOTE procedure synthesized additional samples that resulted in a total sample size of 10293. The following aspects appear to be mentionable:

- Compared to the unbalanced dataset, the results on the hold-out validation set improved substantially with SMOTE by approximately 10.8% for the best two models (gbm by 10.4% from 73.83 to 84.23%, random forests by 11.1% from 73.27 to 84.34%).
- The testing set performance for the best algorithm (gbm) decreased to a small degree of 1.02%, from 73.81 to 72.89%).
- The performance decrease on the testing set was highest for knn (-5.96%) and nnet (-4.66%).

### C. Benchmarking on ADASYN Dataset

Table 5 displays the results of the second benchmarking run, performed on the dataset corrected by ADASYN.

The ADASYN procedure synthesized additional samples that resulted in a total sample size of 10501. The following comments can be made:

TABLE 5: BENCHMARKING RESULTS – ADASYN DATASET

| algorithm | Cross-valid Acc – mean | Cross-valid Acc – sd | Cross-valid Kappa – mean | Cross-valid Kappa – sd | Testingset Accuracy |
|---|---|---|---|---|---|
| gbm | 84.37% | 0.92% | 73.57% | 1.50% | 73.16% |
| rf | 84.62% | 0.93% | 73.75% | 1.58% | 72.51% |
| svmRadial | 78.74% | 0.86% | 62.30% | 1.53% | 68.13% |
| nnet | 77.19% | 0.90% | 59.41% | 1.64% | 66.82% |
| knn | 79.73% | 0.79% | 64.05% | 1.43% | 65.24% |

*Note: Accuracy on 10-fold cross-validation in 10 repetitions, train/test split 85:15*

- Similarly to SMOTE and slightly better, the results on the hold-validation set improved substantially with ADASYN by approximately 11.0% for the best two models (gbm by 10.5% from 73.83 to 84.37%, random forests by 11.4% from 73.27 to 84.62%).
- Inversely to the validation set performance, the testing set performance for the same algorithms was slightly lower than performed on the SMOTE dataset (gbm 73.16% vs. 73.79%, random forests (72.51% vs. 72.79%).
- The standard deviation on the validation set metrics (accuracy and kappa) was slightly higher on the ADASYN dataset than on the SMOTE dataset.

### D. Variable Importance Analysis

The tree-based algorithm implementations, random forests and gbm, provide a ranking of *variable importance*, i.e. a ranking of all features according to their relative influence for the prediction of the target value. The following compares the variable importance between these two algorithms.

Figure 2 shows the visualization of the variable importance for the gbm algorithm. The ranking is largely consistent with the initial ranking when all 91 items were considered. However, the importance values differ, e.g. Q84A has an importance value of 0 whereas it scored 13.6 before.

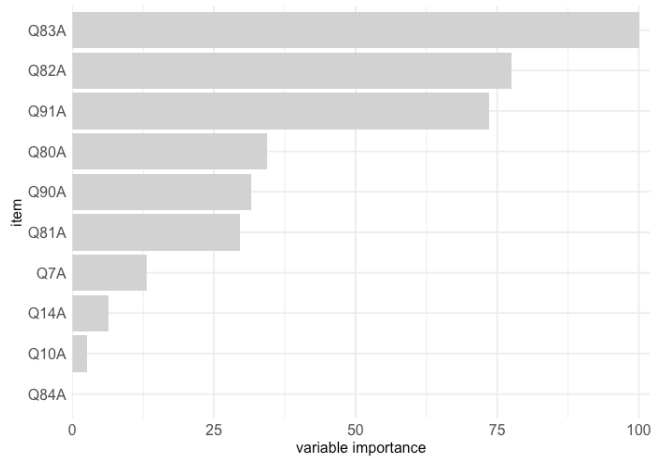

Figure 2: Variable Importance – GBM

The variable importance of random forests shows a completely different picture. As can be seen in Figure 3, random forests acknowledge all ten features in a small range between 75-100.

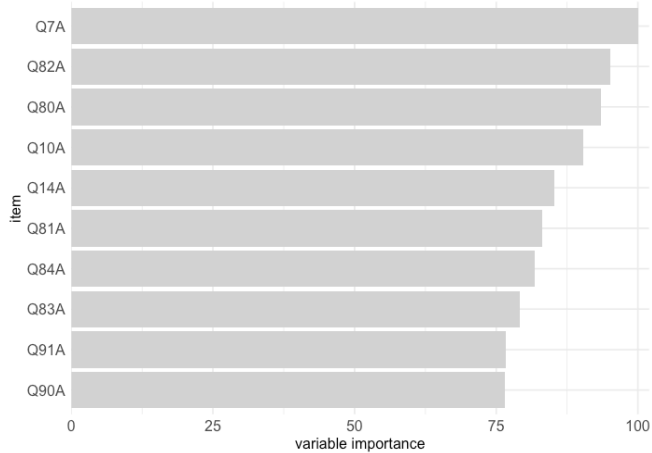

Figure 3: Variable Importance - random forests

This drastic difference illustrates the completely different mechanism between the two algorithms. Random forests grow a high number of decision trees and average their results by majority vote. The frequent reuse of all features (bagging) ensures a representative estimation of their importance independent from random partitioning of the bags. In contrast, gradient boosting machines iteratively grow a single tree and include the error into the next model, boosting the misclassifications in the next iteration. This has proved very efficient but is somewhat more prone to the random selection at the beginning iteration of the algorithm.

## IV. Discussion

The current study contributes to the literature in three ways. First, it predicted people's self-identification as introvert or extrovert by a relatively high accuracy of 73.8% on unseen data. Second, it reduced a measurement instrument from 94 to only 10 features reaching approximately 99% of the accuracy achieved with the full dataset. Third, it applied and compared two class imbalance correction methods, finding ADASYN yielding slightly better results than SMOTE, with a minimal improvement on testing set performance. Fourth, it compared the variable importance ranking between gbm and random forests revealing the difference of the underlying algorithm principles.

ACKNOWLEDGMENT

This research was supported by the Yonsei University Faculty Research Fund of 2019-22-0199.